\documentclass[11pt]{article}
\usepackage{latexsym}
\usepackage{amsmath}
\usepackage{amsfonts}
\usepackage{dsfont}
\usepackage{verbatim}
\usepackage{hyperref}
\usepackage{tikz}
\usetikzlibrary{matrix,arrows}
\usepackage{subcaption}
\usepackage{amsmath,amssymb}
\usepackage{graphicx}
\usepackage{color}
\usepackage{enumerate}
\usepackage{booktabs}
\usepackage{multirow}
\usepackage{url}
\usepackage{enumitem}
\usepackage{notoccite}
\usepackage[ruled,vlined]{algorithm2e}
\usepackage{xcolor,colortbl}
\usepackage{natbib}
\usepackage{authblk}

\title{Exploring UMAP in hybrid models of entropy-based and representativeness sampling for active learning in biomedical segmentation}
\author[1]{Hai Siong Tan\footnote{Corresponding Author. Email: haisiong.tan@pennmedicine.upenn.edu, Phone: 445-2562317.}}
\author[2]{Kuancheng Wang}
\author[1]{Rafe Mcbeth}
\affil[1]{University of Pennsylvania, Perelman School of Medicine, Department of Radiation Oncology, Philadelphia, USA}
\affil[2]{Georgia Institute of Technology, Atlanta, GA, USA}
\date{}                    
\def\be{\begin{equation}}
\def\ee{\end{equation}}
\def\bea{\begin{eqnarray}}
\def\eea{\end{eqnarray}}

\makeatletter
\renewcommand{\@algocf@capt@plain}{above}
\makeatother

\begin{document}

\maketitle

\begin{center}
\textbf{Abstract}
\end{center}
In this work, we study various hybrid models of entropy-based and representativeness sampling techniques in 
the context of active learning in medical segmentation, in particular examining the role of UMAP (Uniform Manifold Approximation and Projection) as a technique for capturing representativeness. 
Although UMAP has been shown viable as a general purpose dimension reduction method
in diverse areas, its role in deep learning-based medical segmentation has yet been extensively explored.  
Using the cardiac and prostate datasets in the Medical
Segmentation Decathlon for validation, 
we found that a novel hybrid combination of Entropy-UMAP sampling technique 
achieved a statistically significant Dice score advantage over the random baseline ($3.2 \%$ for cardiac, $4.5 \%$ for prostate), and 
attained the highest Dice coefficient among the spectrum of 10 distinct active learning methodologies we examined. 
This provides preliminary evidence that there is an interesting
synergy between entropy-based and UMAP methods when the former precedes the latter in a hybrid model of active learning. 
\\[25ex]
\textbf{Keywords}: active learning, biomedical segmentation, UMAP, uncertainty and representativeness sampling
\newpage

\tableofcontents 

\section{Introduction}

\label{sec:intro}

In the biomedical field, segmentation of organs, lesions and other abnormalities in medical imaging is relevant for a wide spectrum 
of clinical applications. For example, in radiotherapy treatment planning, the contours performed by dosimetrists, medical physicists and radiation oncologists are critical in identifying the regions to be irradiated with high dose or those to be spared. Over the years,
the application of deep learning in this area has been shown to be useful in reducing the manual labor involved in annotating 
medical images (see e.g. \citep{Liu} for a recent review), a task of which manual implementation typically involves a large amount of effort and time \citep{daCruz}. Yet a caveat is that training a neural network for performing accurate segmentation generally requires an
enormous amount of contoured images (\citep{Lin,Chen}). 

Active learning (see e.g. \citep{Wolfram,settles} for comprehensive reviews) furnishes a potential framework for optimizing the performance of deep learning models with a minimal amount of training datasets. 
A central problem in active learning frameworks lies in seeking an appropriate sampling technique to iteratively construct an optimal subset of the available data for annotation by physicians, dosimetrists, medical physicists and other human experts. 
Typically, some acquisition function employs
the trained neural network model to pick new training samples from a pool of unlabeled data, which are subsequently annotated
by the human oracle. The newly annotated samples are then included in the training dataset and the neural network model is trained based on this updated dataset. Such a process is iterated till the required performance of the model is reached. 

For biomedical segmentation, active learning has proven to be very useful in reducing the size of datasets required \citep{Less}. 
Two commonly used principles \citep{settles} are those of  \emph{uncertainty sampling} \citep{bilgic,Lewis,Dagan} and \emph{representativeness sampling} \citep{fu2013,ienco,wu2006}. While
the former involves choosing samples with higher Shannon entropy to lower model uncertainty in each iteration, 
the latter involves using a measure of similarity to choose samples most representative of a possibly diverse dataset. 
In recent literature, these measures were often applied to high-dimensional feature vectors derived from the neural network
\citep{Less,Zheng,Yang}, for example, those that emerge as output from the bottleneck layer of a U-Net model. Fusions of these two principles
were often used \citep{Less,Sharma,Nath,Yang,MEAL} and extended to similar frameworks of continual/lifelong learning \citep{van,LwF,Shin,baweja,gonzal}.

Recently in \citep{MEAL}, a novel region-based active learning termed `Manifold Embedding-based Active Learning' was proposed. It
used a combination of entropy-based sampling, and representativeness sampling based on a dimension reduction technique known 
as UMAP (Uniform Manifold Approximation and Projection) developed in \citep{umap}. This technique is a graph-based
nonlinear dimension reduction method of which design is fundamentally motivated by principles of category theory and algebraic topology. Its first use in active learning appeared in \citep{MEAL} which applied this technique successfully to segmentation tasks involving CamVid and Cityscapes datasets. In \citep{MEAL}, the basic idea behind using UMAP was to invoke it to render a low-dimensional representation 
of high-dimensional abstract feature vectors which exist as intermediate quantities in the convolutional neural network (in \citep{MEAL}, DeepLabV3 \citep{Deeplab} was used as the network architecture). Such a low-dimensional projection is then used to represent the image data distribution, upon which the authors of \citep{MEAL} then implemented representativeness sampling. 
From the theoretical perspective, UMAP appears to be potentially relevant for segmentation-based neural network models 
that translate image data distributions to high-dimensional feature vector spaces, since 
UMAP was designed to discover lower-dimensional projections that faithfully preserve topological and geometrical features. This was the primary motivation for
the study in \citep{MEAL}.

To our knowledge, there has been no previous attempt at incorporating
UMAP into some active learning strategy in the context of biomedical segmentation. 
Notably, in the recent work of \citep{yan}, UMAP was found to be rather powerful in providing 
a low-dimensional representation of thalamus tissue signatures (arising from T1-weighted, T2-weighted images, and 
various MRI diffusion measurements) that led to effective
segmentation of thalamic nuclei using a $k$-nearest neighbor algorithm. This suggests that UMAP may enact a useful role 
as a dimension reduction method for representations for biomedical images more generally. 

Motivated by the results of \citep{MEAL} and \citep{yan}, 
here we present a preliminary exploration of how UMAP could potentially fit into an active learning scheme for biomedical image
segmentation. To enable a comparison against a wide variety of active learning methods proposed in the vast literature 
on deep learning-based biomedical segmentation, 
a recent benchmarking framework 
was presented by Burmeister et al.  in \citep{Less} where the two main classes of query methods -- uncertainty and representativeness 
sampling -- were analyzed with respect to openly accessible MRI datasets of the Medical Segmentation Decathlon \citep{Decathlon}. 
Here, we aligned ourselves to several methodological aspects of \citep{Less}, evaluating the query strategies on 
two particular datasets of the \citep{Decathlon} (heart and prostate segmentation) which have been noted for their intra-subject variability,
and adopted a similar model architecture (2D U-Net \citep{olaf}) as in \citep{Decathlon} for a cleaner comparison of results.

We study a spectrum of hybrid models of entropy-based and representativeness sampling involving
UMAP, principal component analysis and random baseline algorithms. As cautiously reviewed in \citep{Less}, past results had revealed that it is still typically difficult for an active learning query strategy to beat the random baseline by a significant margin, despite the number of elaborate proposals presented. 
Thus, in our work where we aim to distinguish between different active learning methods, we harness a suite of evaluation metrics involving various metrics based on voxel overlap (Dice, Precision, Sensitivity, Volumetric Similarity) and Hausdorff distance - related metrics to parametrize the performance of each active learning method holistically. The main novel contributions of our work are as follows. 
\begin{itemize}
\item 
We introduced UMAP as a dimension reduction technique in representativeness sampling as part of active learning for a biomedical segmentation task, and experimented extensively with various hybrid models of entropy-based and representativeness sampling.

\item 
We furnished experimental evidence that when a UMAP-based representativeness sampling was performed after entropy-based sampling, this novel combination of Entropy-UMAP demonstrated statistically significant superiority over the random baseline in terms of Dice scores.  Performance advantages were weaker for two other algorithms involving UMAP, suggesting that there is a potential synergy between entropy and UMAP methods when the former precedes the latter in a hybrid model. 

\item Our novel Entropy-UMAP active learning method led to about $25\%$ and $43\%$ drops in the amount of training samples required
to achieve the maximal Dice scores for the prostate and cardiac datasets respectively, thus affirming the clinical relevance of active learning in a manner consistent with similar work in literature (e.g. \citep{Less} ). Among all the 11 methods considered here, Entropy-UMAP scored the highest Dice score, and in particular, its Dice score superseded the best scores for each dataset as reported in \citep{Less} ($0.957$ vs $0.901$ for cardiac and $0.815$ vs $0.584$ for prostate dataset) with which our work shared similar model architecture and number of epochs per active learning iteration.

\end{itemize}

\section{Methods}
\label{sec:Methods}
In this Section, we provide a more elaborate explanation of the active learning 
methods explored in this work, various implementation details of our methodology, and 
how our approach relates to other active learning-based works in literature.

\subsection{On active learning}
\label{sec:ActiveLearning}
We begin by formulating the essential ideas underpinning active learning. 
Let the entire training dataset be split into a labeled set $\mathcal{D}_L$ and
an unlabeled set $\mathcal{D}_U$. The labeled set is equipped with ground truth labels and
can be used to train the neural network model at any iteration, whereas the unlabeled set does not carry 
ground truth labels. 
Each iteration of active learning involves picking a number ($N_u$)
of images $\tilde{X}$ from the unlabeled set $\mathcal{D}_U$ using some acquisition/query function
$\mathcal{Q}$, 
\be
\label{query}
\{ \tilde{X}_1, \tilde{X}_2, \ldots, \tilde{X}_{N_u} \} = \mathcal{Q} \left(  \{ X: X \in \mathcal{D}_U \} \right),
\ee
which represents a selection criterion such that after sending these unlabeled images to the human expert (e.g. the radiation oncologist annotating the contours) for annotation, they can then be added to the labeled pool for model training.
\be
\mathcal{D}_U  \rightarrow \mathcal{D}_U \setminus   \{ \tilde{X}_1, \ldots, \tilde{X}_{N_u} \}, \,\,\,\,\,\,
\mathcal{D}_L  \rightarrow \mathcal{D}_L  \cup   \{ \tilde{X}_1, \ldots, \tilde{X}_{N_u} \}.
\ee
The acquisition function $\mathcal{Q}$ ranges from simply a random choice to more complicated selection strategies. 
As mentioned in \citep{Less,Wolfram}, two main principles are uncertainty sampling and representativeness sampling. 
For the former, an information-theoretic measure is used to quantify the model uncertainty
associated with each image in $\mathcal{D}_U$, and at each iteration, $N_u$ samples with the largest uncertainty 
are selected. A common measure is the Shannon entropy \citep{shannon} which we adopt in this work. 
Formally, we can express
such a query function as 
\be
\label{Qentropy}
\mathcal{Q}_{entropy} =    \underset{\{ \tilde{X}_1, \ldots \tilde{X}_{N_U}  \} }{ \text{arg max}}\,\,
\sum_{\kappa \in N_{classes}}\,\, \sum_{i \in \tilde{X}}
\left( - P_i^{(\kappa)} \log P_i^{(\kappa)} \right), \qquad P_i = M_{output} \left(\tilde{X}; \vec{W} \right),
\ee
where $\kappa$ is summed over pixel classes, $i$ is the pixel index, and $P_i$ the model's prediction
arising from the activation function of its final output layer, with $\vec{W}$ being its weights. The principle behind 
the uncertainty sampling is that when we feed the model with the data that it is most unsure of, the added samples 
may then inform the model more effectively at each iteration. 

For representativeness sampling, the sampling principle is to capture data points
for a labeled subset that shares similar properties as the entire data distribution, and thus, in this sense, the query function acts as a measure of `representativeness'.  A class of such acquisition functions takes not the 
raw image itself but some feature vector generated by the neural network model as the domain. 
This feature vector is typically high in dimensionality which can be reduced via some technique such as Principal
Component Analysis or UMAP. Formally, we can express the query function as follows. Denoting the 
feature vector by $M_{f} \left( X \right) $,
the reduced dimension by $d$, and the dimension reduction map by $\Phi_{dim}$, we have
\bea
&&\Phi_{dim} : M_{f} \left( X \right) \rightarrow M^{(d)}_{f} \left(  X \right), \cr
\label{Qrep}
&&\mathcal{Q}_{rep} = f\left(  
\tilde{\mathcal{C}}
\right), \,\,\, 
\tilde{\mathcal{C}} =
\underset{\mathcal{C} }{ \text{arg min}}\,\, 
\sum^{N_u}_i \sum_{X \in \mathcal{C}_i }
\lVert   M^{(d)}_f \left( X \right)  - \mu_i
\rVert^2, \,\,\, \mu_i = \frac{1}{\lvert S_i \rvert} \sum_{X \in \mathcal{C}_i} M_f^{(d)} \left( X \right),
\eea
where $\mu_k$ are the representative centroids of $N_u$ number of clusters ($\tilde{\mathcal{C}}_i$)  that can be obtained by performing K-means clustering, and 
$f$ is some function of the cluster set chosen to quantify representativeness. For example, in our work here, $f$ yields a set of $N_u$ samples that are closest
to each centroid of the clustering $\tilde{C}$ in one of our sampling methods \ref{PCA}.

In this work, we also consider hybrid models that leverage upon both uncertainty and representativeness sampling. 
One way of fusing them is to perform each sequentially, effectively adopting a query function that is a composition of both. For example, for the hybrid model of entropy-representativeness sampling, where entropy-based sampling precedes the latter, the query function is 
$\mathcal{Q} = \mathcal{Q}_{rep} \circ\mathcal{Q}_{entropy}$. 
Let $N_c$ be the size of an intermediate set of samples $\mathcal{D}_{int}$ that is 
drawn by the query function, with $N_c > N_u$. We have 
\bea
\mathcal{D}_{int} = \{ X^*_1, \ldots, X^*_{N_c} \} &=& \mathcal{Q}_{entropy} \left(  \{ X: X \in \mathcal{D}_u \} \right), \cr
\label{hybrid}
\{ \tilde{X}_1, \tilde{X}_2, \ldots, \tilde{X}_{N_u} \} &=& \mathcal{Q}_{rep}  \left(  \{ X^*: X^* \in \mathcal{D}_{int} \} \right).
\eea
Conversely, for the hybrid model where representativeness sampling precedes entropy-based sampling, we 
switch the order of $\mathcal{Q}_{entropy}$ and $\mathcal{Q}_{rep}$ above. Equation \eqref{hybrid} describes 
the structure of the hybrid schemes examined in this work.

In Figure \ref{fig:sketch} below, we provide a schematic sketch of our hybrid models of entropy-based and representativeness methods. 
Both methods are anchored on the same U-Net used for prediction, and the dimension reduction techniques of PCA and UMAP are applied on the high-dimensional feature vector of U-Net's bottleneck layer. The budget hyperparameter $N_u$ is generally selected to be much smaller than the size of the complete dataset, and yet not too small to avoid instability during each active learning iteration (see, e.g. \citep{Less} ). For our experiments, we picked these hyperparameters to be $(N_c, N_u) = (24, 16)$ and $(12,6)$ for the cardiac and 
prostate datasets respectively, guided by ablation experiments which showed them to provide a good balance for active learning on these datasets.\footnote{The choice of $N_u = 16$ for cardiac dataset was also adopted in \citep{Less}. }

\begin{figure}[h!]
 \centering
 \includegraphics[width=\textwidth]{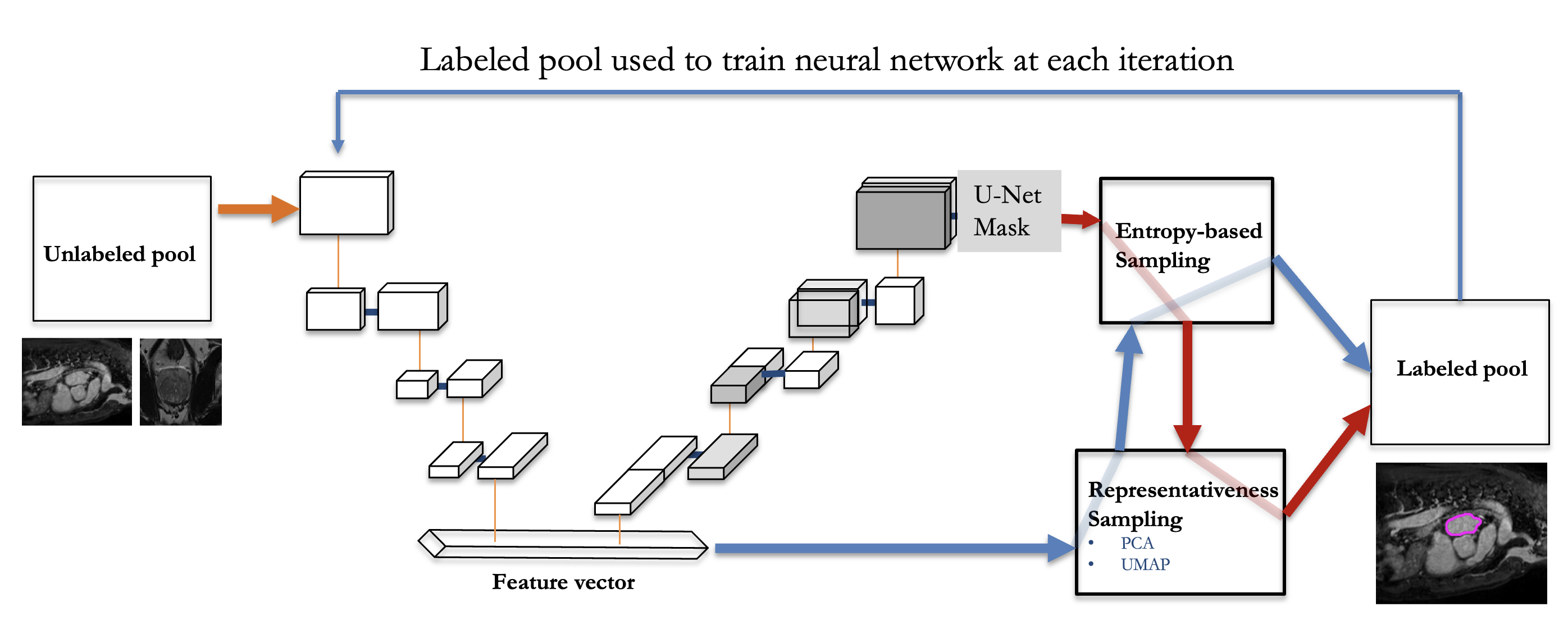}
 \caption{A simple sketch depicting the overall structure and logical flow of our active learning methods. The blue lines above
pertain to `representativeness-Entropy' hybrid algorithms whereas the red ones pertain 
to  the class of `Entropy-representativeness'. We also studied purely entropy-based and representativeness sampling methods for comparison. For the annotation stage prior to being in the labeled pool,
we included the ground truth masks of the images as the equivalent step of a medical expert manually segmenting the images.}
\label{fig:sketch}
\end{figure}

\subsection{Hybrid combinations of entropy-based and representativeness sampling techniques for active learning}
\label{sec:Hybridcombinations}

In the following, we outline the various query strategies adopted in this work in finer details. 
In each active learning iteration, we picked $N_u$ images from the unlabeled pool using a specific sampling protocol, and train the neural network for 10 epochs. Like in \citep{Less}, we fix the total number of active learning iterations to be 
$50$ (or 500 epochs in total).

\begin{enumerate}[label=\textbf{S.\arabic*}]

\item \label{RS} \textbf{Random sampling}: This is the baseline for comparison, and can be implemented with the shortest training time. To our knowledge, in much of past literature, it has been generally noted that there appears to be no universal method that proves to be robustly superior than this baseline, at least for active learning in medical segmentation \citep{Less}. 

\item \label{ES} \textbf{Entropy-based sampling}: We computed the Shannon entropy of each image in the unlabeled pool defined 
as in \eqref{Qentropy}, and picked $N_u$ most uncertain samples from unlabeled pool $\mathcal{D}_U$ for model training.

\item \label{PCA} \textbf{PCA-Representativeness sampling}: The botteneck layer of the U-Net architecture was adopted 
as the defining layer of the feature vector $M_f(X)$  \footnote{In our U-Net model, the bottleneck layer's feature vector's dimensionality is 819,200.} associated with each input image $X$.
We performed a dimension reduction of $M_f(X)$ to a two-dimensional vector $M^{(2)}_f (X)$ using 
Principal Component Analysis \citep{pca}.  
On this auxiliary two-dimensional plane and within the unlabeled pool, we then performed K-means clustering with $N_u$ cluster centers. In each active learning iteration, for each of $N_u$ cluster centroids, the image with its reduced feature vector $M^{(2)}_f (X)$ being closest to the centroid is picked to be labeled. There were thus $N_u$ `representative' images chosen in total.

\item \label{UMAP} \textbf{UMAP-Representativeness sampling}: Like in \ref{PCA}, the feature vector was 
defined via the output of the bottleneck U-Net layer, but this time, we used the method of UMAP \citep{umap} to perform the dimension reduction. K-means clustering with $N_u$ cluster centers was then performed with the $N_u$ images closest to each cluster centroid being  picked to enter the labeled pool. The implementation of UMAP brought with it the choice of several hyperparameters. For reading convenience, we relegate their discussion to Appendix A, where more technical details are explained.

\item \label{ER} \textbf{Entropy-Random sampling}: In this hybrid method, we first performed entropy-based 
sampling based on \eqref{Qentropy}, but choosing $N_c > N_u$ most uncertain images instead of $N_u$. 
Out of this set, we then randomly picked $N_u$ to enter the labeled pool, furnishing a simple way to choose 
a high-entropy set that may potentially capture more of the image data distribution.

\item \label{EUMAP} \textbf{Entropy-UMAP sampling}: As in \ref{ER}, we first identified $N_c$  most uncertain images
according to \eqref{Qentropy}, forming the intermediate pool $\mathcal{D}_{int}$  \eqref{hybrid}. 
Like in \ref{UMAP}, we reduced the dimensionality of the feature vector space of images in $\mathcal{D}_{int}$ to just two. Then we used K-means clustering to identify 
$N_u$ cluster centroids. For each element of $\mathcal{D}_{int}$, we computed its cluster label.
Finally, from a random starting label, we iterated through the $N_c$ labels, each time randomly choosing an element that carried the same label. Now, $\mathcal{D}_{int}$ may not carry all of the distinct $N_u$ labels, in which case we repeated this loop until we obtained $N_u$ samples. A pseudocode is provided in Appendix B for additional clarity.

\item \label{EPCA} \textbf{Entropy-PCA sampling}: This is very similar to \ref{EUMAP} except that instead of 
using UMAP for dimension reduction of the feature vector space, we used PCA instead.

\item \label{RE} \textbf{Random-Entropy}: This is like the reverse of \ref{ER}. Now instead of the high entropy-subset, we picked a representative subset of size $N_c$ before choosing $N_u$ most uncertain of them according to \eqref{Qentropy}. 

\item \label{PCAE}  \textbf{PCA-Entropy}: This is like the reverse of \ref{EPCA}. We first picked $N_c$ images identical to the procedure in 
\ref{PCA}, and then chose $N_u$ most uncertain of them according to \eqref{Qentropy}. 

\item \label{UMAPE}  \textbf{UMAP-Entropy}: This is like the reverse of \ref{EUMAP}. We first picked $N_c$ images following the procedure in 
\ref{UMAP}, and then chose $N_u$ most uncertain of them according to \eqref{Qentropy}.

\end{enumerate}

The query strategies \ref{ER}, \ref{EUMAP}, \ref{EPCA} represent various
forms of the `entropy-representativeness' sampling principle whereas \ref{RE}, \ref{UMAPE}, \ref{PCAE}
are examples of `representativeness-entropy' sampling algorithms.

\subsection{Implementation details: datasets, neural network and evaluation metrics}
\label{sec:OnDatasets}

Open-source datasets were employed for our validation of active learning strategies. 
In this work, we chose to use two MRI image datasets drawn from the Medical Segmentation Decathlon \citep{Decathlon}: cardiac (left atrium) mono-modal MRI  and prostate (combining central gland and peripheral zone) MRI (T2 and ADC modes), both of which were 
described in \citep{Decathlon} to display large inter-subject variability. The ground truth images were annotated by medical experts in 
Radboud University, Nijmegen Medical Centre for the prostate dataset and those in King's College London for cardiac dataset \citep{Decathlon}.

For our neural network architecture, for a cleaner comparison of final results, we 
aligned with the work in \citep{Less} (which is devoted to a benchmarking framework for active learning in biomedical segmentation), adopting
a standard 2D U-Net with a 4-layer structure for both 
upsampling and downsampling paths. In \citep{Less}, it was noted that they preferred 2D over 3D U-Net models \citep{iek}
as they found that the latter did not lead to performance gap while the former enjoyed higher training speeds and was found to be more suitable as active learning model architectures.
The segmentation mask was recovered after passing the output through a sigmoidal activation layer. For the decoder, we picked bilinear interpolation as the unpooling method, whereas max-pooling was adopted in the encoder. Each layer was equipped with one $3\times 3$ convolutional filter, and the total number of trainable parameters of our U-Net was of the order $\sim 10^6$. After some experimentation, we find the learning rate of $10^{-4}$ to be the appropriate order-of-magnitude, consistent with the figure reported in \citep{Less}. For loss function,
we used minor variants of the Dice loss depending on what we found to be most effective and stable for each dataset: the Dice-BCE loss 
was adopted for the cardiac dataset and the focal Dice loss (with $\gamma = 3$ in eqn. \ref{focal}) was used for the prostate dataset. Explicitly, 
denoting the ground truth label and model's prediction for pixel $i$ by $y^{i}_t, y^i_p$ respectively, these loss functions are defined as
\bea
\label{focal}
\mathcal{L}_{focal} &=&  1 - \left( 
\frac{2  \frac{1}{N} \sum^N_{i=1}  y^i_t y^i_p  }{ \frac{1}{N} \sum^N_{i=1}  (y^i_t + y^i_p)}
\right)^\gamma, \\
\label{dice}
\mathcal{L}_{Dice-BCE} &=& 1 -  \left( 
\frac{2  \frac{1}{N} \sum^N_{i=1}  y^i_t y^i_p  }{ \frac{1}{N} \sum^N_{i=1}  (y^i_t + y^i_p)}
\right) - \frac{1}{2} \sum_{i=1}^N \left[  y_t^i \log y^i_p + (1- y^i_t)\log (1-y^i_p )  \right],
\eea
where $N$ is the total number of pixels for a single 2D image, and we took the mean of the loss functions over all training image samples. 
We also set a common initial condition for the various active learning methods by first training the U-Net with $10\%$ of the entire dataset picked uniformly at random until a training Dice accuracy of about $10 \%$ was attained.\footnote{When the initial labeled set $\mathcal{D}_L$ is small, it has been shown that random selection is generally
considered as a good baseline for initializing active learning (e.g. \citep{Provost,Mittal}). } We checked that equipped with such an initial condition,
the model did not suffer from any severe cold start-like problems such as instability in learning curves during the initial round. 
Each active learning query strategy was then implemented on this baseline neural network.  

Hyperparameters such as those of the loss function were first selected with a 4:1 split of the training dataset, excluding a separate hold-out set that comprises of $20\%$ of the entire dataset. This independent hold-out set was then used to report the model prediction accuracies in Section \ref{sec:Results}.
All experiments 
were implemented in Tensorflow with Adam Optimizer on a NVIDIA A100 GPU. For random sampling, the time taken for 50 iterations was about 30 min and 2 hrs for the prostate and cardiac datasets. Pure representativeness/entropy-based sampling took $\sim 30 \%$ more time, while hybrid models took twice the time. This trend was observed for each of the two datasets.

In most works discussing active learning methodology in medical segmentation,
the Dice coefficient is typically employed as the measure of performance.
 Here, 
with the goal of furnishing a more holistic assessment of the query strategies,
we 
kept track of a spectrum of evaluation metrics that are sensitive not only to the amount of voxel matching
but also to the geometry and topology of the contours. 
For voxel-overlap based metrics, we chose (i)Dice (ii)Precision (iii)Sensitivity (iv)volumetric similarity. 
For measuring geometrical deviations of the contours, we chose (i)mean surface distance (ii)average Hausdorff distance
(iii)Hausdorff distance at $95^{\text{th}}$ percentile.

\subsection{Some comments on related works}
\label{sec:comments}
In this Section, we furnish some discussion on a number of intricate relations connecting between our work and related papers in literature. We would first like to point out
some crucial aspects by which our Entropy-UMAP sampling
principle defined in \ref{EUMAP} depart from the query 
method presented in \cite{MEAL}, even when the general idea 
of first performing uncertainty sampling followed by 
representativeness sampling is shared. 

In \cite{MEAL}, there were two proposals for how UMAP can be
used for dimension reduction. The first one (that 
the authors called `MEAL') proceeds by implementing the UMAP
algorithm on the \emph{entire dataset} prior to any active learning and this UMAP embedding also remains preserved throughout, unaffected by how the neural networks' weights change in each active learning epoch. The UMAP embedding is a dimension-reduction of the feature vector space that belongs to a
pre-trained model which, specifically in \cite{MEAL}, was
a MobileNetV2 \cite{sandler} backbone trained on the ImageNet
dataset \cite{ImageNet}. The model architecture itself was taken
to be DeepLabV3 \cite{chen2017} which augments the backbone
by atrous convolutions (among other techniques). Presumably,
this feature vector distribution and its UMAP embedding were
still faithful descriptions of the underlying data distribution
so that a further clustering on this space can capture representativeness. It should also be noted that the dataset
supporting the pre-trained model was characterized by 1000 classes whereas the examined datasets (CamVid and Cityscapes) had
just 30 classes each. It is not clear to us whether/how the faithfulness of the UMAP embedding (to the original image data distribution) may be compromised, especially when it is no longer updated during active learning iterations where the model weights -- especially those that define the feature vector space -- are refined during network training.

In contrast, our Entropy-UMAP method defined in \ref{EUMAP}
involves implementing UMAP as a dimension reduction of the 
feature vector space of the neural network itself (not some
pre-trained model exposed to a distinct dataset). Since with every active learning iteration the model was updated, the UMAP embedding was also updated at every iteration capturing the changes in the feature vector space that arise during model training. This adaptive nature of our implementation of UMAP 
lies in contrast to the MEAL method of \cite{MEAL}. As explained
earlier, we employed the UMAP embedding of the feature vector space
defined by the model and the \emph{unlabeled} dataset $\mathcal{D}_u$,
rather than the entire dataset. Our acquisition functions were thus sensitive purely to the
largest possible set that we were querying from at each iteration. We picked the eventual query set to the one equipped with the maximal number of distinct cluster labels associated with the K-means clustering implemented on the unlabeled dataset $\mathcal{D}_u$. In contrast, for the MEAL method, the final query set is the set of centroids (or operationally, samples closest to centroids) associated with the K-means clustering performed only on the much smaller high entropy subset rather than
entire dataset or $\mathcal{D}_u$. The accompanying notion of representativeness is thus \emph{restricted to only the high entropy subset for the MEAL method}. This intermediate subset was taken to be much smaller in size than the entire dataset, being less than $\sim 1\%$ of the full training dataset for both our work
and theirs \cite{MEAL}.

Now in \cite{MEAL}, the authors also considered a variant which they called `MEAL-FT' where the low-dimensional representation
was the UMAP embedding \emph{not} of the entire dataset but only of
the high uncertainty subset. Presumably, this was still based on the
feature vector space arising at some point in architecture of the MobileNetV2 backbone. Another significant difference was that
in our case, we took the embedding of the full unlabeled set rather than the high uncertainty subset. The former was a much larger set
than the latter $\mathcal{D}_{int}$ which was a very small proportion of the data distribution. 
Fundamentally, these differences characterize the distinct ways by which \emph{representativeness} is implicitly defined in the active learning scheme.

Let us now formalize our preceding explanations explicitly.
For our Entropy-UMAP approach, the overall form of the 
acquisition function reads as
\bea
\mathcal{Q}_{entropy-umap} &=& \mathcal{Q}_{rep} \circ \mathcal{Q}_{entropy}
\left( \{X : X \in \mathcal{D}_u \} \right),
,\,\,\,
\mathcal{Q}_{rep} = \underset{\{ \tilde{X}_1, \ldots \tilde{X}_{N_u} \in \mathcal{D}_{int}  \} }{ \text{arg max}}\,\,\lvert
\{  \tilde{\mathcal{C}} \}_{\neq \,}
\rvert   \cr
\label{eumapE}
\mathcal{D}_{int} &=& \{ X^*_1, \ldots, X^*_{N_c} \} = \mathcal{Q}_{entropy} \left(  \{ X: X \in \mathcal{D}_u \} \right),
\eea
where $X^*, \tilde{X}, X$ denote samples in the high-entropy intermediate set, query set and the unlabeled set respectively,
and $\{\tilde{\mathcal{C}} \}_{\neq \,}$
denotes the set of unique cluster labels, the clustering being
that of K-means implemented on the UMAP embedding
of $\mathcal{D}_u$, i.e. 
\bea
&&\tilde{\mathcal{C}} =
\underset{\mathcal{C} }{ \text{arg min}}\,\, 
\sum^{N_u}_i \sum_{X \in \mathcal{C}_i }
\lVert   M^{(d)}_{umap} \left( X \right)  - \mu_i
\rVert^2, \,\,  \mu_i = \frac{1}{\lvert S_i \rvert} \sum_{X \in \mathcal{C}_i} M_{umap}^{(d)} \left( X \right), \,\, X \in \mathcal{D}_u  \cr
\label{cluster1}
&&M^{(d)}_{umap} : F(\mathcal{D}_{u} ) \rightarrow \mathbb{R}^d,
\eea
where $F(\mathcal{D}_{u})$ is the
feature vector space of samples
in the unlabeled set, and in our work here, we took $d=2$. We also note that 
for $\mathcal{Q}_{rep}$ in \eqref{eumapE},
we picked the final query set to be a set endowed with the maximal number of distinct cluster labels as a measure of
representativeness defined with respect to 
the unlabeled set $\mathcal{D}_u$. 

On the other hand, for the methods of 
\cite{MEAL}, the UMAP embedding is defined with respect to different subsets. Explicitly, 
\bea
\label{meal1}
M^{(\text{MEAL})}_{umap} &:& F^{\text{(pretrained)}}
\left(  \mathcal{D}_L \cup \mathcal{D}_U  \right) \rightarrow 
\mathbb{R}^2, \\
\label{meal2}
M^{(\text{MEAL-FT})}_{umap} &:& F^{\text{(pretrained)}}
\left(  \mathcal{D}_{int}  \right) \rightarrow 
\mathbb{R}^2,
\eea
where $F^{\text{(pretrained)}}$ is the feature vector space arising from a certain layer in their
MobileNetV2 backbone \cite{MEAL} that is pre-trained 
on the ImageNet \cite{ImageNet} dataset. 
For the MEAL algorithm, the choice in \eqref{meal1} implies that 
the UMAP projection is an approximation of 
the underlying feature vector space of the model obtained by just using the pre-trained backbone. For the MEAL-FT algorithm, the choice in \eqref{meal2}
implies that representativeness is only defined with respect to the high-entropy subset $\mathcal{D}_{int}$
which is much smaller than the initial unlabeled 
dataset. 

Also, for both MEAL and MEAL-FT algorithms,
$\mathcal{Q}_{rep}$ in \eqref{eumapE} yields the $N_u$ cluster centroids $\mu_i$
defined in \eqref{cluster1} with the clustering
performed only on $X \in \mathcal{D}_{int}$ (instead 
of $X \in \mathcal{D}_u$ in our case). This implies
that for the MEAL method, the notion of representativeness inherited a `mixed' origin since
the UMAP embedding was defined for a 
feature vector space belonging to $\mathcal{D}_L \cup \mathcal{D}_U$ yet the clustering was performed assuming a much smaller $\mathcal{D}_{int}$ as the entire space of data distribution. For the MEAL-FT method, the nature of both clustering and the UMAP projection was consistent like in our method, yet the defining data distribution (and its associated notion of representativeness) was restricted to just
$\mathcal{D}_{int}$.

In this work, we have chosen to focus on a set of experiments that probe the
efficacy of various hybrid models defined via \eqref{hybrid}. Across the recent literature, while
details of how we implement the fusion of uncertainty and representativeness sampling may differ, the
organizing principles are nonetheless similar in spirit. In \citep{Nath}, where datasets from the Medical 
Segmentation Decathlon \citep{Decathlon} were also used, it was found that increasing the frequency
of `uncertain data' (defined similarly via \eqref{Qentropy} ) was a useful active learning strategy.
The authors of \citep{Nath} also proposed a balance between uncertainty sampling and some method to capture representativeness of the training dataset. Instead of the query function composition we invoked in \eqref{hybrid}, they adopted a single query function that computes the difference between \eqref{Qentropy}
and a variant of \eqref{Qrep} that is based on the mutual information between $\mathcal{D}_L$ and 
$\mathcal{D}_U$ which was intended to extend the diversity of the sampled data in each iteration beyond just uncertain ones. It was emphasized in \citep{Nath} that their results suggested that the efficacy of any specific active learning method is probably data-dependent, rather than being universal in nature. 

In \citep{Sharma}, where active learning was applied to brain tumor lesion segmentation from MR images,
the authors used a combination of uncertainty and representativeness sampling in the same sense as \citep{Nath} by defining a query function that measures the difference between uncertainty and representativeness scores. 
Compared to \citep{Nath}, their measure of representativeness itself was much closer by definition to ours, 
with K-means clustering being applied and the cluster centroids acting as representative points of the data distribution. 

The form of our hybrid acquisition function in \eqref{hybrid} appears to be closest to the work in \citep{Yang} (see 
in particular Section 2.3 of \citep{Yang}), where the authors argued that `uncertainty is a more important
criterion' and thus, a subset of uncertain images was first picked before extracting a final, smaller subset
of data that carried the largest representativeness. A difference is that in \citep{Yang}, for each feature vector, 
the cosine-similarity between two feature vectors was used to measure their degree of similarity, and that
the representativeness query function $\mathcal{Q}_{rep}$ in \eqref{Qrep} is defined by taking samples
that maximizes the cosine similarity between $\mathcal{D}_{int}$ and $\mathcal{D}_U$. 
Validating their approach on gland and lymph node segmentation , their results showed that active learning can reduce the training dataset by about 50$\%$ which is rather similar in magnitude to our results for the cardiac dataset.

Most recently in \cite{shao1}, a very interesting active learning algorithm (termed as SSDR-AL) was presented which could be interpreted as a hybrid model synthesizing both uncertainty and representativeness sampling. In 
\cite{shao1}, the context was point cloud semantic segmentation with the experimental datasets 
being two point cloud benchmarks (S3DIS \cite{s3dis} and Semantic3D \cite{semantic3D}).
A graph-based reasoning provided the organizing principle for merging the two active learning methods, the nodes of the graph being the samples with high uncertainty and the edges defined by invoking both Euclidean and Chamfer distances as measures of separation. A graph
aggregation operation was then invoked to project points into a diversity space where representative samples were identified via farthest point sampling. We observe that in contrast to our methodology, two different notions of distance measures were involved in quantifying representativeness. It would be interesting to explore the role of UMAP in this active learning strategy by, for example, using an UMAP embedding of the feature space instead of the feature space itself in SSDR-AL.

A commonly cited SOTA active learning method is the work of \citep{coreset} which formulated the problem of active learning as a core-set problem and
showed that to a good approximation, it is equivalent to a $k$-center problem for which, using our terminologies, 
the query function can be
effectively written as 
\be
\mathcal{Q}_{\text{core-set}} =  \underset{\{ \tilde{X}_1, \ldots  \ldots \tilde{X}_{N_U}  \in \mathcal{D}_U \} }{ \text{arg max}}\,\,
\underset{X \in \mathcal{D}_L}{\text{min}}
\lVert   \tilde{X} - X 
\rVert^2.
\ee
Notably, this method departs from using feature vectors as representative objects of the image distribution and from using
any form of uncertainty measures. For completeness, we applied the $k$-center greedy algorithm described in \citep{coreset}
as an active learning strategy for our datasets, so that the hybrid models we constructed could also be compared against this method.

Finally, we note that in our work and all the above-mentioned
literature on active learning, each iteration of model training selects
a subset of the training dataset but does not involve any form of data augmentation. Recently in \cite{shao2}, the authors proposed a novel data augmentation method that adds to the training dataset new virtual training samples pairing foregrounds with various backgrounds replacing the original ones. This `counterfactual learning paradigm' was shown to be able to disentangle
foreground and background more effectively and thus ameliorate the biased activation problem
in the context of object localization \cite{shao2}, at least for the CUB-200-2011 \cite{CUB} and ImageNet \cite{ImageNet} datasets. To our knowledge, this interesting data augmentation method has yet been incorporated into active learning. It would be interesting to see if this `counterfactual representation synthesis' can be invoked as part of some form of weakly-supervised sampling principle in active learning, in particular for patch-based segmentation of medical images.

\section{Results}
\label{sec:Results}

At the end of the 50${}^\text{th}$ iteration, the mean Dice scores of all the active learning methods were all close to and statistically 
indistinguishable from the Dice score distribution of the model trained on the entire dataset from scratch. In this 
aspect, Entropy-UMAP yielded the highest scores for both datasets : $0.96$ for the cardiac dataset and $0.82$ for the prostate dataset (vs $0.96$ and $0.78$ for the model trained on the complete dataset). The validation Dice learning curves appeared to begin to 
saturate at the end of the 50${}^\text{th}$ iteration for both datasets and all active learning methods, where the labeled training dataset
is only about $57\%$ and $75\%$ of the complete training datasets for the cardiac and prostate cases respectively.

\begin{figure}[h]
\centering
\begin{subfigure}{.53\textwidth}
  \centering
  \includegraphics[width=\linewidth]{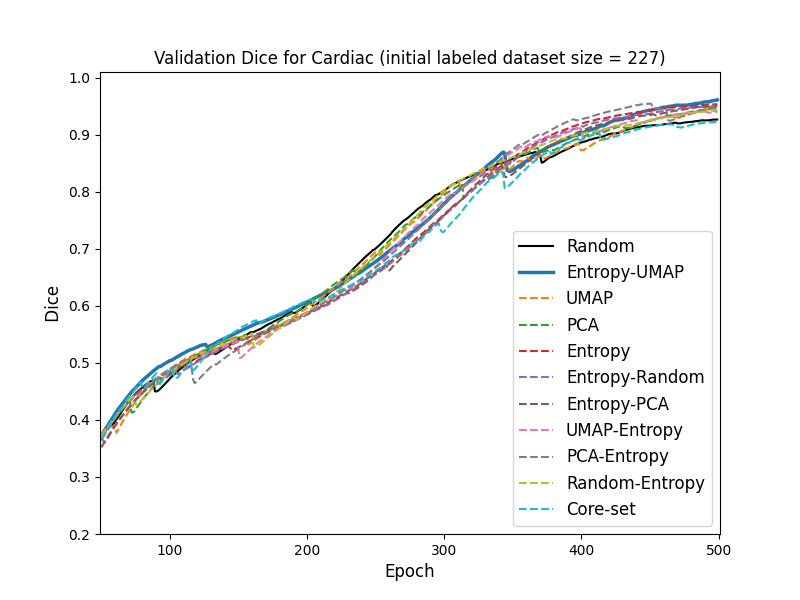}
\end{subfigure}%
\begin{subfigure}{.53\textwidth}
  \centering
  \includegraphics[width=\linewidth]{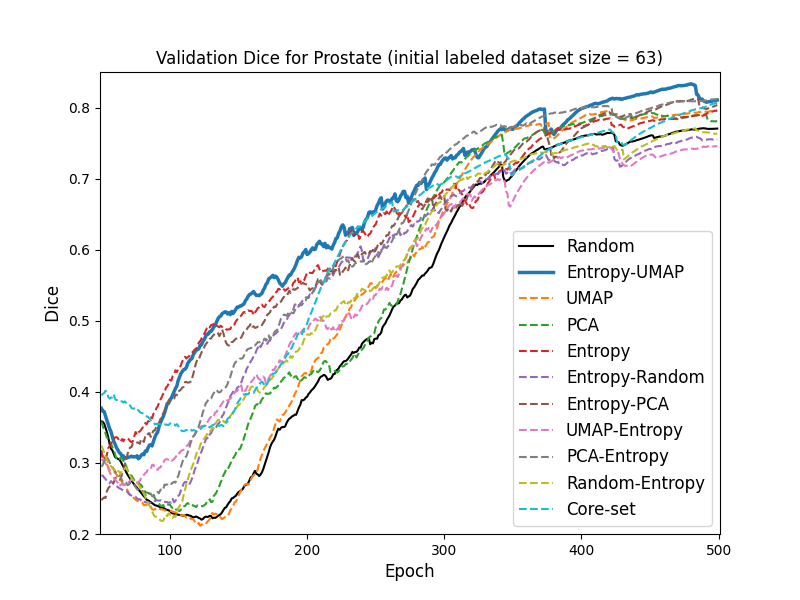}
\end{subfigure}
\caption{We display the evolution of the validation Dice scores for various hybrid models of uncertainty and representativeness sampling. Each active learning iteration involved 10 training epochs and acquired 6 and 16 samples for the prostate and cardiac datasets respectively. }
\label{fig:learningcurves}
\end{figure}

Each active learning model was initiated with 63 and 227 labeled training samples for the prostate and cardiac datasets respectively, and every iteration 
involved 10 epochs, and acquired 6 and 16 samples for the prostate and cardiac datasets respectively.
In Figure \ref{fig:learningcurves}, we plot the evolution of the validation Dice scores for various hybrid models of uncertainty and representativeness sampling.
For the cardiac dataset, starting from about 40${}^\text{th}$ iteration, Dice accuracy for the various methods began to supersede that of the random baseline of which performance was leading the group of methods from the 25${}^\text{th}$  to the 40${}^\text{th}$  iteration.  
 For the prostate dataset, there were more evident fluctuations and variation among the various methods. Transient brief dips in the validation Dice scores were
also present for some methods up till about the 13${}^\text{th}$ iteration where all active learning scheme recovered from this cold-start-like instability. 
Following \cite{Less}, we ended active learning at the  
50${}^\text{th}$ iteration, at which point the mean Dice scores of all methods were statistically 
indistinguishable from the Dice score distribution of the model conventionally trained on the entire dataset. By this point, all learning curves
appeared to begin to level off.

\begin{figure}[h]
\centering
\begin{subfigure}{.53\textwidth}
  \centering
  \includegraphics[width=\linewidth]{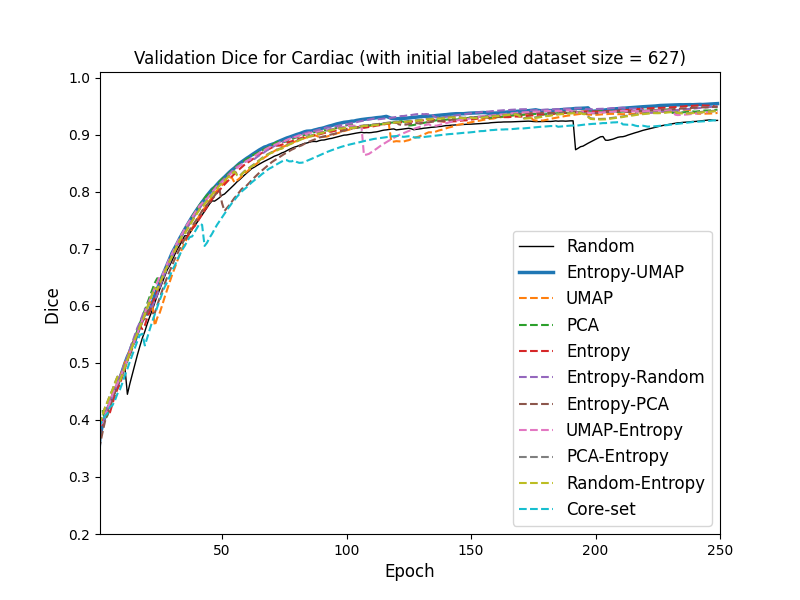}
\end{subfigure}%
\begin{subfigure}{.53\textwidth}
  \centering
  \includegraphics[width=\linewidth]{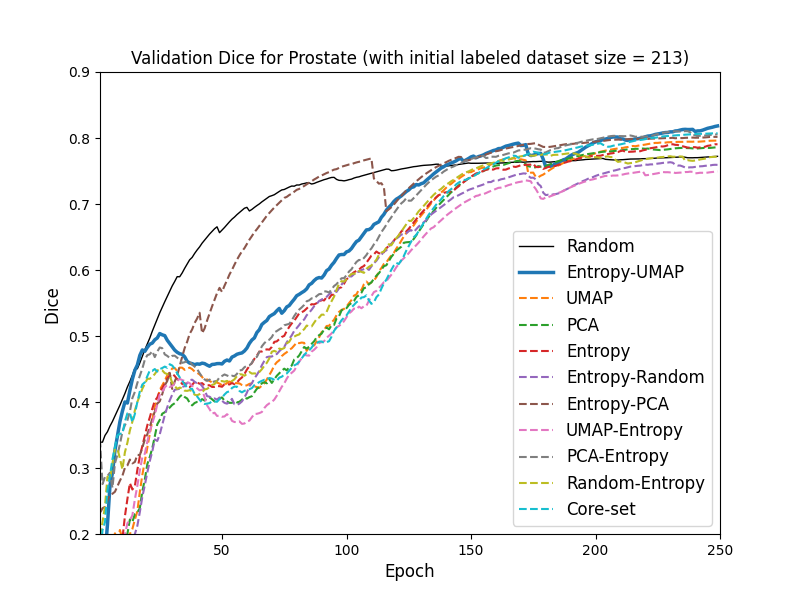}
\end{subfigure}
\caption{Here, we display the learning curves for larger initial (labeled) datasets, with the initial size of each $\mathcal{D}_L$ picked such that the number of samples acquired per iteration and final size of $\mathcal{D}_L$ remained the same
as in Fig. \ref{fig:learningcurves} while halving the number of active learning iterations.}
\label{fig:learningcurvesR}
\end{figure}

Although we kept to a low proportion of the training dataset for the initial number of labeled samples
(similar to other related works, e.g. \cite{Less,Yang,coreset}) for our primary analysis in this paper, 
in Figure \ref{fig:learningcurvesR}, we plot the learning curves for the same set of models but each trained with 
larger initial labeled dataset $\mathcal{D}_L$. This allowed us to probe whether the training evolution dynamics retained the same qualitative features when initial $|\mathcal{D}_L|$ was enlarged. Our choice of parameters was mainly motivated by the corresponding choice in \cite{MEAL} where the authors picked a larger initial $\mathcal{D}_L$ such that the final number of acquired patches remained identical while the number of active learning epochs was reduced by roughly half. Although we didn't find statistically significant differences in final Dice scores relative to the cases with smaller initial $|\mathcal{D}_L |$, we found that for the prostate-based models, the transient dips in validation Dice scores in early training stages were absent for the random sampling method while they persisted for the rest. For the cardiac case, the learning rate appeared to stabilize and gradually approach an asymptote starting from about the 100${}^\text{th}$ epoch. Overall, these comparisons suggested that 
the main qualitative features of the 
learning curves were preserved for most methods for our choices of initial
$|\mathcal{D}_L |$, the more evident difference being that the models trained with a larger initial $|\mathcal{D}_L |$ converged slightly more quickly towards the end. For the rest of the paper, we focused on the setting with the smaller $|\mathcal{D}_L |$ depicted in Figure \ref{fig:learningcurves}, which resembled more closely the typical initial conditions for active learning at least in our context of medical segmentation (e.g. \cite{Less, Yang, Sharma}).

For the MRI datasets chosen in our work, each carries a significant inter- and intra-subject variability. Against this backdrop, 
we performed various (one-tailed) statistical tests to indicate the significance for the observed differences in the mean evaluation scores relative
to the random baseline. Scores with associated p-value $<.05$ are indicated in bold. 
Paired two-sample t-test was performed for the voxel overlap-based metrics (Dice, Precision, Sensitivity) and 
unpaired two-sample t-test was indicated for Hausdorff distance-based ones (average Hausdorff, mean surface distance, 
95 $\%$ Hausdorff). These metrics were evaluated on images where the neural network 
model predicted only single contours, and thus on a subset of validation dataset which differed slightly among the 
active learning methods.  For the metric of volumetric similarity where the typical score distribution was found to be so skewed such that the typical median is about $40 \%$ higher than the mean, we performed the Wilcoxon signed-rank test for a comparison of medians across the 
set of active learning methods. 
In Tables \ref{table:1} and \ref{table:2}, we collect various mean metric scores for each active learning method with their one-sided p-values in brackets. 

\noindent
(i) \emph{cardiac dataset}: 
\\[1ex]
In terms of Dice coefficient, all active learning methods superseded the random baseline with small but statistically significant margins, 
with the best-performing one being Entropy-UMAP attaining a $3.2 \%$ improvement over the random baseline. 
For the three Hausdorff distance-based metrics (average Hausdorff, mean surface distance and $95\%$ Hausdorff),
Entropy-UMAP also attained the best scores which were $\sim 14-17 \%$ smaller than those of the random baseline. 
The superiority of UMAP-entropy over the random baseline was relatively less evident and for pure UMAP, it was better than the baseline only
in terms of Dice score. 

Apart from Entropy-UMAP, two other active learning algorithms which demonstrated superiority over the random baseline 
in all evaluation metrics (with $p<.05$) were Entropy and Entropy-PCA. We note that the pure representativeness sampling methods were not distinguishable from the random baseline. Compared to pure Entropy, adding Random sampling preceding or following Entropy appeared to
produce a markedly inferior performance.  The Core-set method demonstrated statistically significant superiority over the random baseline only in the aspect of Sensitivity where it scored the highest among all active learning methods. 
\\[2ex]
(ii) \emph{prostate dataset}: 
\\[1ex]
In contrast to the cardiac MRI case, there were much less areas showing statistically significant margins of difference 
from the random baseline. For Dice and Precision, Entropy-UMAP displayed the highest and statistically significant improvement of
about $5 \%$ over the baseline, whereas for Sensitivity,  Entropy-PCA was the method that stood out among all others. 
For the Hausdorff distance-based metrics, the methods of Entropy, Entropy-UMAP, Entropy-Random, Entropy-PCA and PCA-Entropy displayed better (lower) mean scores over the baseline for all three metrics, though in terms of the overall score distribution, these 
improvements over the baseline were statistically insignificant.  The Core-set method demonstrated statistically significant superiority over the random baseline in the areas of Dice score and Sensitivity.

\begin{table}[htbp]
\small
\setlength{\tabcolsep}{1.5pt}
{%
\begin{tabular}{|p{1.7cm}|p{1.1cm}|p{1.2cm}|p{1.2cm}|p{1.2cm}|p{1.2cm}|p{1.2cm}|p{1.2cm}|p{1.2cm}|p{1.2cm}|p{1.2cm}|p{1.2cm}|} 
\hline
$\,$ & R & E & P & U & EU & ER & EP& UE & PE & RE & CS\\
\hline
\textit{Average} & 0.941 & \cellcolor{lightgray!40}  \textbf{0.795} & 0.902 & 1.16 &  \cellcolor{lightgray!40} \textbf{0.777} & 1.32 
& \cellcolor{lightgray!40}  \textbf{0.800} & \cellcolor{lightgray!40} \textbf{0.880}  & \cellcolor{lightgray!40}  \textbf{0.835} & 0.987 & 0.927 \\
\textit{Hausdorff} & $\,$ &\cellcolor{lightgray!40}  $p<.05$ & $p=.13$  & $p=.07$  & \cellcolor{lightgray!40} $p<.05$  
& $p=.24$  & \cellcolor{lightgray!40}  $p<.05$  & \cellcolor{lightgray!40}  $p<.05$  &  \cellcolor{lightgray!40} $ p<.05$  & $p=.37$ &$p=.34$  \\
\hline
\textit{Mean} & 0.951 & \cellcolor{lightgray!40}  \textbf{0.803} & 0.910 & 1.18 & \cellcolor{lightgray!40} \textbf{0.785 }& 1.33 
& \cellcolor{lightgray!40}  \textbf{0.807} & \cellcolor{lightgray!40}  \textbf{0.888} & \cellcolor{lightgray!40}  \textbf{0.848} & 0.997  &0.935 \\
\textit{S.D.} & $\,$ &\cellcolor{lightgray!40}  $p<.05$ & $p=.12$  & $p=.07$  & \cellcolor{lightgray!40} $p<.05$  
& $p=.24$  & \cellcolor{lightgray!40}  $p<.05$  & \cellcolor{lightgray!40}  $p<.05$  &  \cellcolor{lightgray!40} $ p<.05$  & $p=.37$ & $p=.33$  \\
\hline
$95\%$ & 2.70 & \cellcolor{lightgray!40}  \textbf{2.41} & 2.60 & 2.99 &  \cellcolor{lightgray!40} \textbf{2.32 }& 2.86 
& \cellcolor{lightgray!40}  \textbf{2.40} & 2.53 & \cellcolor{lightgray!40}  \textbf{2.46} & 2.71 &2.53 \\
\textit{Hausdorff} & $\,$ &\cellcolor{lightgray!40}  $p<.05$ & $p=.19$  & $p=.09$  & \cellcolor{lightgray!40} $p<.05$  
& $p=.39$  & \cellcolor{lightgray!40}  $p<.05$  &  $p=.07$  &  \cellcolor{lightgray!40} $ p<.05$  & $p=.47$  & $p=.06$ \\
\hline
\textit{Dice} & 0.925 & \cellcolor{lightgray!40}  \textbf{0.952} &  \cellcolor{lightgray!40}  \textbf{0.944} &  \cellcolor{lightgray!40}  \textbf{0.942} &  \cellcolor{lightgray!40} \textbf{0.957 }&   \cellcolor{lightgray!40} \textbf{0.949}
& \cellcolor{lightgray!40}  \textbf{0.949} &  \cellcolor{lightgray!40}  \textbf{0.940} & \cellcolor{lightgray!40}  \textbf{0.943} &  \cellcolor{lightgray!40}  \textbf{0.941} & 0.921 \\
$\,$ & $\,$ &\cellcolor{lightgray!40}  $p<.05$ &   \cellcolor{lightgray!40} $p<.05$  & \cellcolor{lightgray!40}  $p<.05$  & \cellcolor{lightgray!40} $p<.05$  
&  \cellcolor{lightgray!40}  $p<.05$  & \cellcolor{lightgray!40}  $p<.05$  &  \cellcolor{lightgray!40}  $p<.05$  &  \cellcolor{lightgray!40} $ p<.05$  &  \cellcolor{lightgray!40}  $p<.05$ & $p=.36$   \\
\hline
\textit{Precision} & 0.946 & \cellcolor{lightgray!40}  \textbf{0.960} &  \cellcolor{lightgray!40} \textbf{0.961} & 0.956 &  \cellcolor{lightgray!40} \textbf{0.964 }& 0.954 
& \cellcolor{lightgray!40}  \textbf{0.958} &  \cellcolor{lightgray!40}  \textbf{0.959} & 0.958 & 0.952  & \cellcolor{lightgray!40} \textbf{0.912}\\
$\,$ & $\,$ &\cellcolor{lightgray!40}  $p<.05$ &  \cellcolor{lightgray!40} $p<.05$  & $p=.10$  & \cellcolor{lightgray!40} $p<.05$  
& $p=.17$  & \cellcolor{lightgray!40}  $p<.05$  &   \cellcolor{lightgray!40} $p<.05$  &  $ p=.06$  & $p=.22$ &  \cellcolor{lightgray!40} $p<.05$  \\
\hline
\textit{Sensitivity} & 0.946 & \cellcolor{lightgray!40}  \textbf{0.968} & 0.952 & 0.954 &  \cellcolor{lightgray!40} \textbf{0.969}&  \cellcolor{lightgray!40}  \textbf{0.969}
& \cellcolor{lightgray!40}  \textbf{0.967} & 0.950 & 0.958 & 0.955 &  \cellcolor{lightgray!40} \textbf{0.980}\\
$\,$ & $\,$ &\cellcolor{lightgray!40}  $p<.05$ & $p=.23$  & $p=.24$  & \cellcolor{lightgray!40} $p<.05$  
&  \cellcolor{lightgray!40}   $p<.05$  & \cellcolor{lightgray!40}  $p<.05$  &  $p=.34$  &  $p=.11$  & $p=.19$ &  \cellcolor{lightgray!40} $p<.05$  \\
\hline
\textit{Volumetric } & 0.946 & \cellcolor{lightgray!40}  \textbf{0.950} & 0.936 & 0.948 &  \cellcolor{lightgray!40} \textbf{0.972}&  \cellcolor{lightgray!40}  \textbf{0.959}
& \cellcolor{lightgray!40}  \textbf{0.956} & 0.933 & \cellcolor{lightgray!40}  \textbf{0.948} & 0.937  & 0.949\\
$Similarity$ & $\,$ &\cellcolor{lightgray!40}  $p<.05$ & $p=.60$  & $p=.50$  & \cellcolor{lightgray!40} $p<.05$  
&  \cellcolor{lightgray!40} $p<.05$  & \cellcolor{lightgray!40}  $p<.05$  &  $p=.19$  &  \cellcolor{lightgray!40} $ p<.05$  & $p=.93$  & $p=.99$ \\
\hline
\end{tabular}
}
\caption{Table collecting mean metric scores (median for Volumetric Similarity only) for each active learning strategy implemented on the cardiac MRI dataset.  Bold values
with shaded cells pertain to those with $p<.05$. \textbf{Abbreviations for active learning methods}: R=Random \ref{RS}, E=Entropy \ref{ES}, P=PCA \ref{PCA},  U=UMAP \ref{UMAP}, EU=Entropy-UMAP \ref{EUMAP}, ER=Entropy-Random \ref{ER}, EP=Entropy-PCA \ref{EPCA}, UE=UMAP-Entropy \ref{UMAPE}, PE=PCA-Entropy \ref{PCAE}, RE=Random-Entropy \ref{RE}, CS=Core-set \cite{coreset}.}
\label{table:1}
\end{table}

\begin{table}[h!]
\small
\setlength{\tabcolsep}{1.5pt}
{%
\begin{tabular}{|p{1.7cm}|p{1.1cm}|p{1.2cm}|p{1.2cm}|p{1.2cm}|p{1.2cm}|p{1.2cm}|p{1.2cm}|p{1.2cm}|p{1.2cm}|p{1.2cm}|p{1.2cm}|} 
\hline
$\,$ & R & E & P & U & EU & ER & EP& UE & PE & RE & CS\\
\hline
\textit{Average} & 3.71 & 3.53 & 3.66 & 3.85 &  3.48 & 3.43
& 3.25 & 4.59  & 3.51 & 3.86  & 3.617\\
\textit{Hausdorff} & $\,$ &  $p=.31$ & $p=.46$  & $p=.40$  &  $p=.27$  
& $p=.23$  &  $p=.12$  &  $p=.17$  & $ p=.30$  & $p=.40$  & $p=.41$  \\
\hline
\textit{Mean} & 3.79 & 3.67 & 3.82 & 3.95 & 3.52 & 3.47 & 3.30 & 4.76 & 3.57  & 4.01 & 3.84 \\
\textit{Surf. Dist.} &  $\,$ & $p=.39$ & $p=.47$ & $p=.40$ & $p=.25$ & $p=.21$ & $p=.11$ & $p=.17$ & $p=.29$ & $p=.38$& $p=.46$   \\
\hline
$95\%$ &  9.10 & 8.58 & 9.09 & 9.61 & 8.43 & 8.09 & 8.10 & 9.89 & 8.77 & 9.23 & 9.47   \\
\textit{Hausdorff} & $\,$  & $p=.27$ & $p=.50$ & $p=.32$ & $p=.21$ & $p=.11$ & $p=.12$ & $p=.29$ & $p=.36$ & $p=.46$ & $p=.35$  \\
\hline
\textit{Dice} &  0.770 & 0.791 & 0.786 & 0.794 & \cellcolor{lightgray!40} \textbf{0.815} & 0.756 & \cellcolor{lightgray!40}   \textbf{0.804} 
& 0.744 & \cellcolor{lightgray!40} \textbf{0.810} & 0.766  & \cellcolor{lightgray!40}  \textbf{0.807} \\
$\,$ & $\,$ & $p=.16$ & $p=.24$ & $p=.16$ & \cellcolor{lightgray!40} $p<.05$ & $p=.22$ & \cellcolor{lightgray!40} $p<.05$ 
& $p=.18$ & \cellcolor{lightgray!40} $p<.05$ & $p=.39$   & \cellcolor{lightgray!40} $p<.05$  \\
\hline
\textit{Precision} & 0.820 & 0.837 & 0.829 & 0.823 & \cellcolor{lightgray!40} \textbf{0.869} & 0.807 & 0.836 
& 0.795 & 0.848 & 0.794 & 0.847 \\
$\,$ & $\,$ & $p=.21$ & $p=.34$ & $p=.45$ & \cellcolor{lightgray!40} $p<.05$ & $p=.26$ & $p=.21$ 
& $p=.21$ & $p=.10$ & $p=.10$ & $p=.13$ \\
\hline
\textit{Sensitivity} & 0.875 & 0.874 & 0.881 & 0.887 &  0.863 & 0.869 & \cellcolor{lightgray!40} \textbf{0.890} 
& \cellcolor{lightgray!40} \textbf{0.858} & 0.879 & 0.882  & \cellcolor{lightgray!40}  \textbf{0.888}\\
$\,$ & $\,$ & $p=.47$ & $p=.27$ & $p=.17$ & $p=.07$ & $p=.22$ &  \cellcolor{lightgray!40} $p<.05$ 
& \cellcolor{lightgray!40} $p<.05$ & $p=.38$ & $p=.21$ & \cellcolor{lightgray!40} $p<.05$  \\
\hline
\textit{Volumetric} & 0.914 & 0.915 & 0.919 & 0.917 & \cellcolor{lightgray!40} \textbf{0.905} & \cellcolor{lightgray!40} \textbf{0.890} & 0.926
& 0.914 & 0.924 & 0.912 & 0.909  \\
$Similarity$ & $\,$ & $p=.91$ & $p=.87$ & $p=.66$ & \cellcolor{lightgray!40} $p<.05$ & \cellcolor{lightgray!40} $p<.05$ &  $p=.36$ 
& $p=.16$ & $p=.78$ & $p=.45$ & p =.28 \\
\hline
\end{tabular}
}
\caption{Table collecting mean metric scores for the prostate MRI dataset. 
In contrast to the cardiac MRI case, there were much less areas showing statistically significant margins of difference 
from the random baseline. }\label{table:2}%
\end{table}

\section{Discussion}
\label{sec:Discussion}

In this paper, 
we introduced UMAP as a dimension reduction technique in representativeness sampling for a couple of
biomedical segmentation tasks, 
and examined its effectiveness in 
various hybrid models of entropy-based and representativeness-based active learning methods. 
Twenty-two experiments based on a 2D U-Net as the core framework were performed using the cardiac and prostate MRI datasets in 
the Medical Segmentation Decathlon \citep{Decathlon}. 
Against the backdrop of strong inter- and intra-subject
variability in the score distribution, we invoked various statistical tests to quantify the significance for the observed differences in the
metric scores relative to the random baseline. Following \citep{Less}, we ran each active learning scheme for 50 iterations/500 epochs at the 
end of which the annotated training dataset comprised of $0.57$ and $0.75$ of the full training set 
for the cardiac and prostate datasets respectively. In terms of Dice scores, 
the best performing method turned out to be Entropy-UMAP which attained Dice coefficients for both datasets similar to those achieved by the model being trained on the entire dataset.

In the following, we summarize the essential traits distinguishing among
the active learning methods in terms of results common in \emph{both} MRI datasets. 
\begin{enumerate}[label=(\roman*)]

\item Entropy-UMAP and Entropy-PCA methods stood out as the 
best performers for both datasets, showing holistic improvements over the random baseline in most evaluation metrics. 
In particular, Entropy-UMAP yielded the best Dice and Precision accuracy scores, and was also superior in Hausdorff 
distance-based 
measures (though statistical significance in this aspect was present only for the cardiac dataset).  

\item The only cases of statistically significant 
improvements of mean scores over the baseline for \emph{both} datasets were observed for these following methods:
\begin{itemize}
\item Dice: Entropy-UMAP, Entropy-PCA, PCA-Entropy
\item Precision: Entropy-UMAP 
\item Sensitivity: Entropy-PCA, Core-set
\end{itemize}
There were no cases of consistent and statistically significant degradation from the random baseline for any method
in terms of any of the seven evaluation metrics. 

\item Pure representativeness sampling (PCA, UMAP) and random sampling 
were generally inferior to entropy-based methods. Relative to their Entropy-Representativeness counterparts, Representativeness-Entropy hybrids were generally weaker in performance 
across the evaluation metrics we used, and for both datasets.
\end{enumerate}

Collectively, our results indicated that when a UMAP-based representativeness sampling is performed after entropy-based sampling, this novel combination of Entropy-UMAP displayed the most evident and statistically significant Dice score advantage over the random baseline for each dataset we examined  ($3.2 \%$ for cardiac, $4.5 \%$ for prostate).  Among all the 11 active learning strategies considered here, Entropy-UMAP scored the highest Dice score, and in particular, its Dice score superseded the best scores for each dataset as reported in \citep{Less} ($0.957$ vs $0.901$ for cardiac and $0.815$ vs $0.584$ for prostate dataset) with which our work shared similar model architecture and number of epochs per active learning iteration.  
Performance advantages were weaker for pure UMAP and UMAP-entropy, suggesting that there is a potential synergy between entropy and UMAP methods when the former precedes the latter in a hybrid model. For the class of hybrid models we constructed in this work via
\eqref{hybrid}, performing uncertainty sampling before representativeness sampling led to a generally superior performance, incidentally realizing the assumption made in an earlier related work of \citep{Yang}. 

A caveat is that there are context-dependent variations of each technique's performance observed here, compatible with that reported in \citep{Nath}, and which suggest that attributes of the underlying image distribution affect the relative efficacies of the sampling principles. 
Here we briefly discuss the potential role of UMAP in domain adaptation techniques which may help us ameliorate context dependence, leaving 
detailed explorations for future work. 
Recall that fundamentally, UMAP furnishes an embedding through a graph construction of the higher-dimensional space that induces a lower-dimensional projection inheriting the topological properties of the original data distribution. Domain adaptation techniques which rely on extracting information from feature vectors can thus potentially benefit from their lower-dimensional projections via UMAP. Among other reasons, this can be attributed to excess inert information being filtered through the projection, and/or enhanced computational efficiency via working in an effective lower-dimensional space. 
An exemplifying set of domain adaptation techniques introduced in the context of segmentation is the recent series of works in 
\cite{Luo3,Luo2,Luo1} which addresses specifically unsupervised domain adaptation
through novel adversarial networks
\cite{Luo1,Luo3} and an `adversarial style mining' approach \cite{Luo2}. Feature vectors 
turned out to be crucial elements in these works, for example, in defining the `adversarial loss functions' of \cite{Luo3}, the feature extractor modules of the neural network model in \cite{Luo1}, etc. 
Notably, t-distributed stochastic neighbor embedding (t-SNE) \cite{tSNE} was used to furnish 2D representations that enabled clearer visualizations of the differences among the various model performances (see e.g. Fig. 4 and 7 of \cite{Luo2}). It would thus be interesting to explore if a dimension-reduction method like UMAP can be used within these adversarial-type models to furnish lower-dimensional projections of feature vectors to be used in place of the original ones, and ultimately whether this can enhance domain adaptation performance beyond providing visualizations of model capabilities.

We hope that the methodologies and results presented here will inspire more 
in-depth exploration of the role of UMAP in active learning strategies, and a better scrutiny of the relation between data distribution
and sampling techniques. Entropy-UMAP sampling method potentially furnishes a basis for an efficient and adaptive active learning strategy for biomedical segmentation that leverages upon the UMAP technique to capture representativeness, ultimately attaining good segmentation performance with minimally available data.


\section*{Declaration of competing interest}
All authors have no conflict of interest to declare.

\appendix

\section{Visualizing UMAP embeddings}\label{apd:geometry}

\begin{figure}[h!]
\centering
\begin{subfigure}{.50\textwidth}
  \centering
  \includegraphics[width=\linewidth]{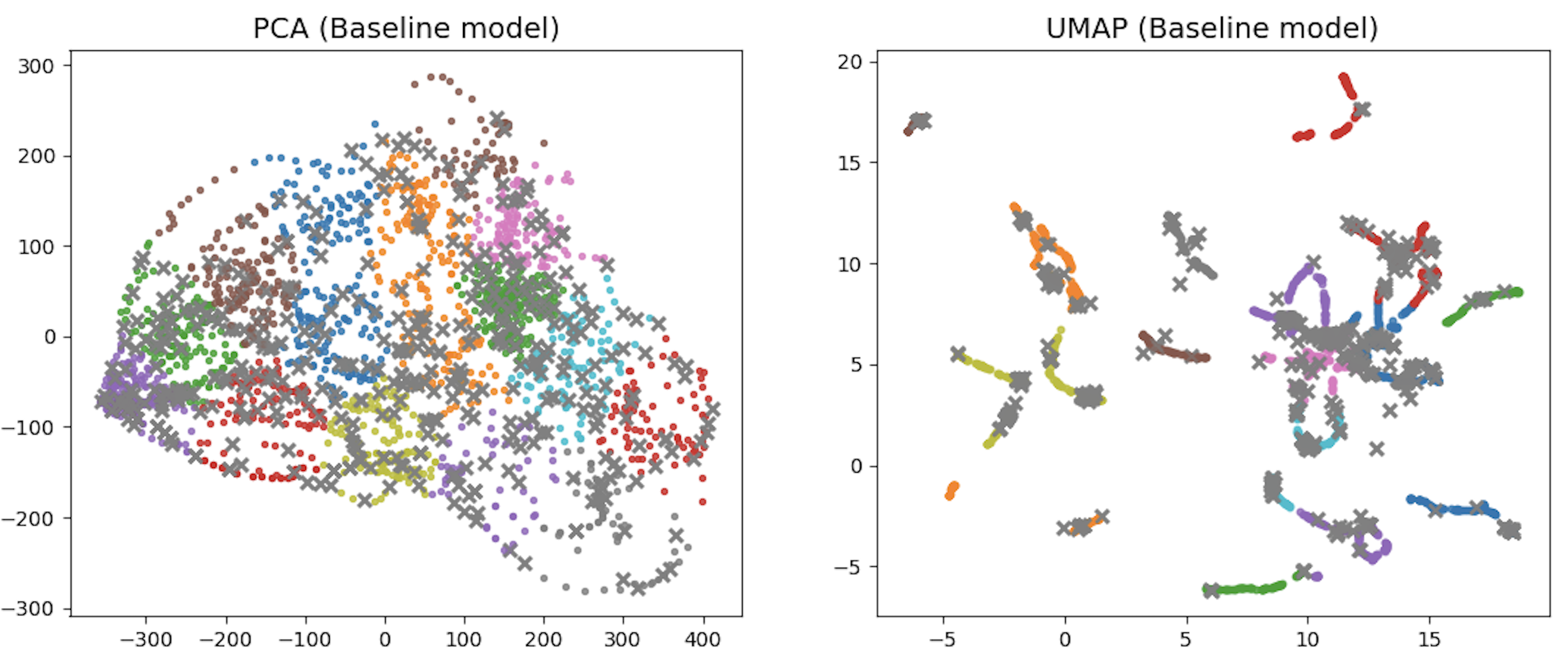}
  \caption{Projected cardiac dataset (baseline model)  }
  \label{fig:PCA}
\end{subfigure}%
\begin{subfigure}{.50\textwidth}
  \centering
  \includegraphics[width=\linewidth]{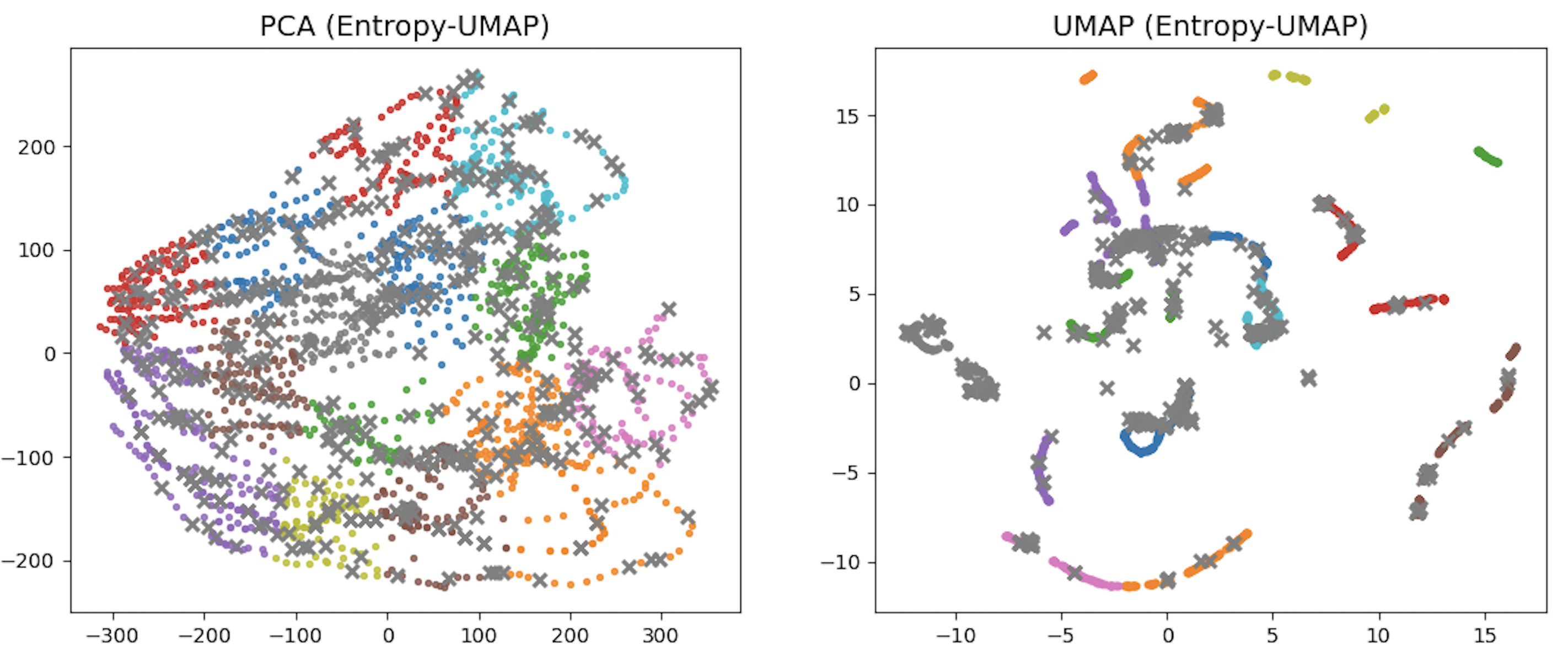}
  \caption{Projected cardiac dataset (trained model)}
  \label{fig:UMAP1}
\end{subfigure}
\begin{subfigure}{.49\textwidth}
  \centering
  \includegraphics[width=\linewidth]{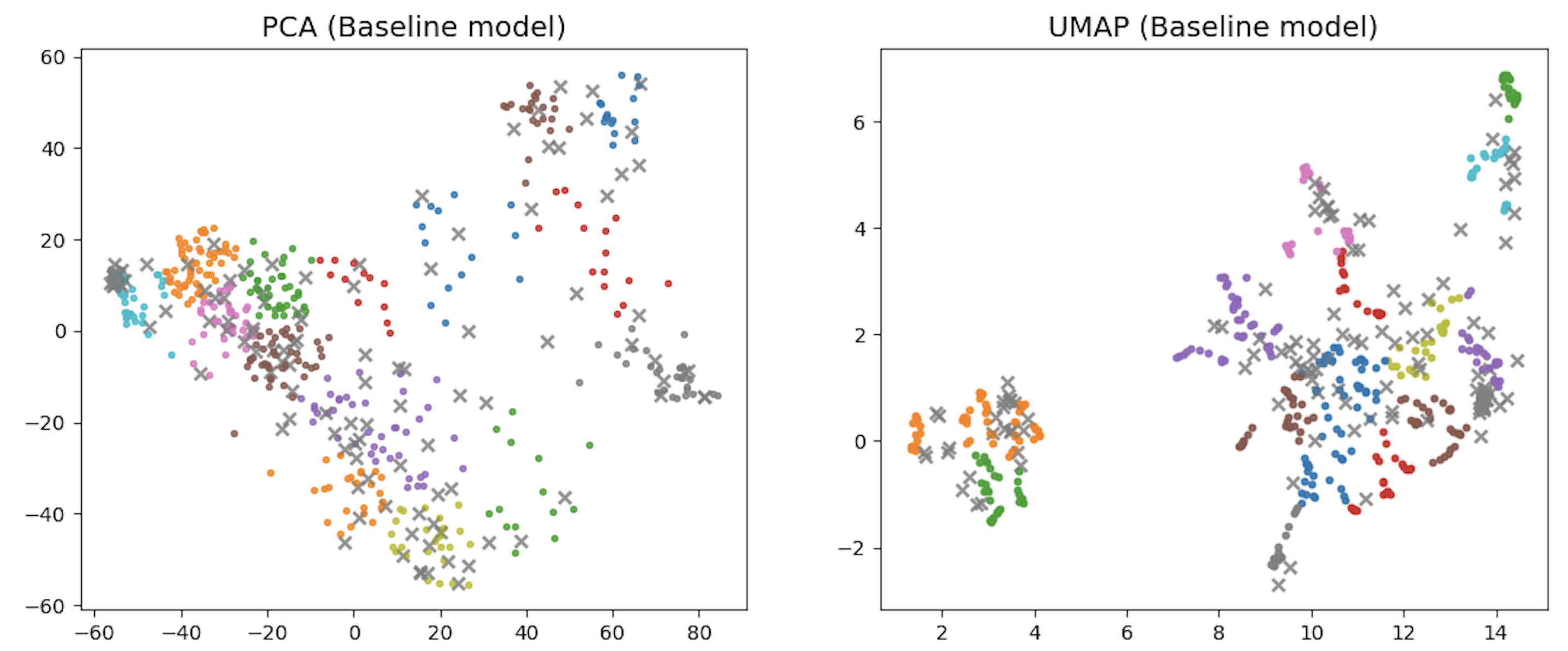}
  \caption{Projected prostate dataset (baseline model) }
  \label{fig:UMAP2}
\end{subfigure}
\begin{subfigure}{.49\textwidth}
  \centering
  \includegraphics[width=\linewidth]{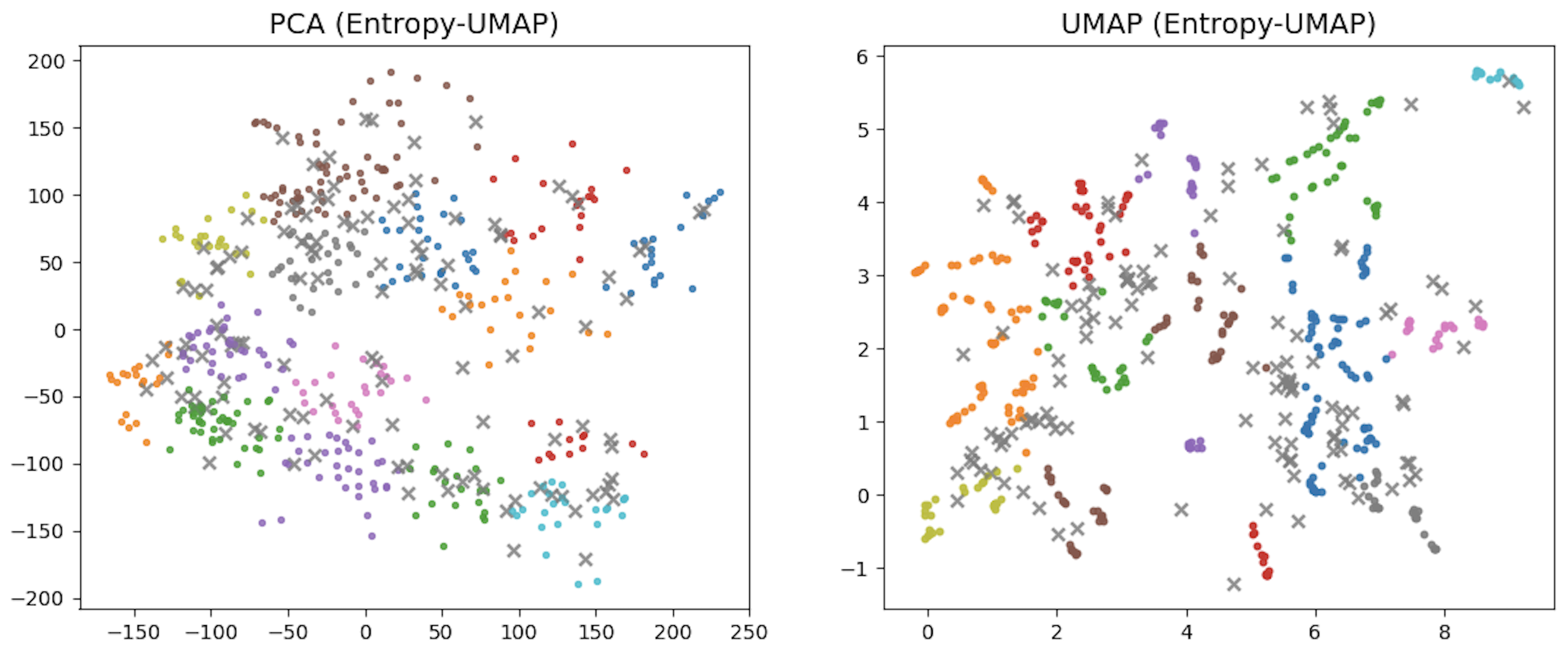}
  \caption{Projected prostate dataset (trained model) }
  \label{fig:UMAP3}
\end{subfigure}
\caption{These are two-dimensional projections of the feature vector space obtained via PCA and UMAP. Grey crosses superimposed on the training dataset correspond to images of the validation dataset transformed according to the PCA and UMAP algorithms that are trained on the training dataset.  For
both PCA and UMAP, the validation data distribution closely aligns with that of the training dataset. The UMAP-transformed space consists of more isolated and linear clusters, in  contrast to the spherical and uniform ones in PCA-transformed space.
 }
\label{fig:umapplots}
\end{figure}

To visualize the two-dimensional projections of the feature vector space obtained in UMAP and PCA, 
in Figure \ref{fig:umapplots} , we plot the PCA-transformed and UMAP-transformed two-dimensional spaces.
The baseline model is the initial model trained on $10 \%$ of the entire dataset, whereas the Entropy-UMAP model refers to the one trained for 50 iterations of Entropy-UMAP active learning method. The original feature vector space is a 819200-dimensional space
obtained from the bottleneck layer of the U-Net.  We picked the following default hyperparameter values indicated in the original UMAP software package \citep{umap} :
\begin{itemize}
\item Number of neighbors = 15. This is the default value in the UMAP package which we also found to be optimum after performing silhouette score analysis for the clustering done in the UMAP-transformed feature vector space. 
In principle, this parameter affects the balance between local and global structure of the image data distribution. 
\item Minimum distance = 0.1. This is the default value which we followed. Lower values tend to lead to clumpier embeddings and this parameter controls how tightly points are packed together in the low-dimensional representation. 
\item Number of components = 2. This is the resulting dimension of the UMAP-transformed feature vector space. We picked it to be identical to the one for PCA-based methods. It is also the default value that is most convenient for data visualization, and was also choice in \citep{MEAL} which implemented active learning for the CamVid and Cityscapes dataset with UMAP as a technique for measuring representativeness.
\item Metric = Euclidean. We found that this default choice also led generally to much faster computations. 
\end{itemize}
A more technical review of how these parameters precisely affect the UMAP algorithm can be found in Section 3.2 of \citep{yingfan}
apart from the original work in \citep{umap}. In Figure \ref{fig:umapplots}, grey crosses superimposed on the training dataset correspond to images of the validation dataset transformed according to the PCA and UMAP algorithms that are trained on the training dataset.  For
both PCA and UMAP, the validation data distribution closely aligns with that of the training dataset. Certain geometrical characteristics of the PCA and UMAP embeddings appear to be independent of the stage of active learning, as well as the type of MRI dataset. The UMAP-transformed space consists of more isolated and curve segment-like clusters, in contrast to the more uniform-distribution-like ones in PCA-transformed space, with these distinctions appearing to be stronger for the cardiac dataset. 

\newpage

\section{Pseudocode for Entropy-UMAP}\label{apd:pseudocode}
In the following, we display an explicit pseudocode for the Entropy-UMAP method as an illustrative example. 
Across our 
twenty experiments, we set the number of active learning iterations $N_K = 50$, the initial labeled
training dataset $D_0$ comprising of $10\%$ of the entire dataset. 
\begin{algorithm}[h!]
	{\caption{Entropy-UMAP pseudocode}}
{%
	\textbf{Initialize}: Unlabeled pool $D_{U}$, Number of active learning iterations $N_K$, Number of query patches $N_u$, 
, Initial labeled pool $D_{0}$, Labeled pool $D_{L}$, 
Train the segmentation model $f_{unet}$ till it attains $\sim 10\%$ validation Dice accuracy. \\
   	\For{$k \gets 1$ \KwTo $K$}{	
  \textbf{Query Procedure:} \\    
 \textbf{Step 0}:  \\
Compute feature vectors of all images  in $D_U$ that arise from the bottleneck layer of $f_{unet }$ with 
		images in the unlabeled pool $D_U$ \;
        $F_{j} = f_{unet}( I_j),\, j = 1,2, \ldots \text{dim}(D_U)$ \\
\textbf{Step 1}: \\  
Train a UMAP $\Phi$ using all feature vectors $F_j$ \;
Set $G = \varnothing$
 \\
Compute UMAP embeddings of the feature vector as \\
		\qquad $G_{j} \gets \Phi(F_{j})$ \\
       		 \qquad $G \gets G \cup G_{j}$\;
Run K-Means clustering on all embeddings $G$ to obtain $N_u$ cluster centroids \\
	    \textbf{Step 2} : \\ Compute Shannon entropy $H_{j}$ of image $I_{j}$ in $D_{U}$ using probabilities
$P_i$ predicted by $f_{unet}$ and using
		$H_{j} =- \sum_{i \in I_j}   P_{i} \log P_{i} + (1- P_{i} ) \log (1- P_{i} ) $ \\
		Select $N_{c}$ images with highest entropy $H_{j}$ to form a high-entropy pool $D_E$ \\
	   \textbf{Step 3} :  \\
		Compute cluster label set $C_E$ which contains cluster labels of all $N_c$ images in the high-entropy pool $D_E$ \\
 	\textbf{Step 4} :  \\
	We now construct the query pool $D_Q$ starting with $D_Q = \varnothing$\;
	$m \gets 0$\;
        \While{$m < N_u$}{
		\For{$k \gets 1$ \KwTo $N_u$}{		
			\If{$k \in C_E$}{
       				Randomly pick an element $I_e$ of the high-entropy pool $D_E$ which has cluster label $=k$ to be an element of the 
				query set $D_Q$ \;
				$m \gets m+1$\;
  				$D_Q \gets D_Q \cup I_e$\;
        	       			$D_E \gets D_E \setminus I_e$, $C_E \gets C_E \setminus$ (cluster of $I_e$)
				}		
		  }   
          }
		\textbf{Update:} $D_{U} \gets D_{U} - D_{Q}$, $D_{L} \gets D_{L} \cup D_{Q}$ \\
		\textbf{Step 5}: Train the model $f_{\theta}$ on $D_{L}$ 
	 } 	
	\Return Final trained model $f_{\theta}$
}%
\end{algorithm}

\newpage

\section{Examples of contoured images}\label{apd:gallery}
\begin{figure}[htbp]
 \centering
 \includegraphics[width=0.9\textwidth]{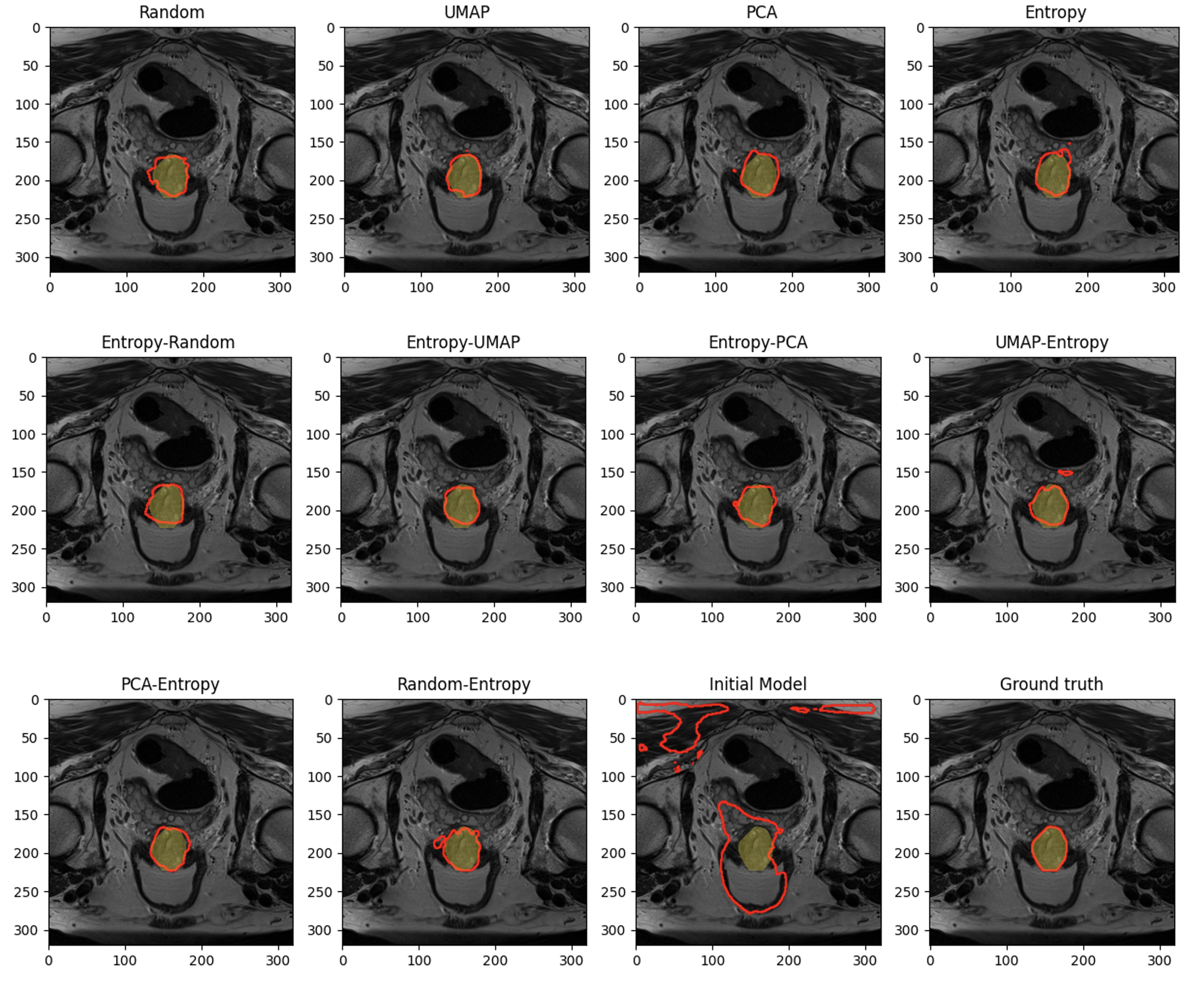}
 \caption{An illustrative set of prostate MRI images with the contours outlined in red for various hybrid or singular models, overlaid on the ground truth mask in yellow which portrays the combined peripheral and central zones. 
These contours were obtained at the 50${}^{\text{th}}$ iteration.
The last two figures in the bottom row correspond to the contours predicted by the initial model before active learning was started and the ground truth.}
 \label{fig:prostateR}  
\end{figure}

\begin{figure}[h]
\centering
\includegraphics[width=\textwidth]{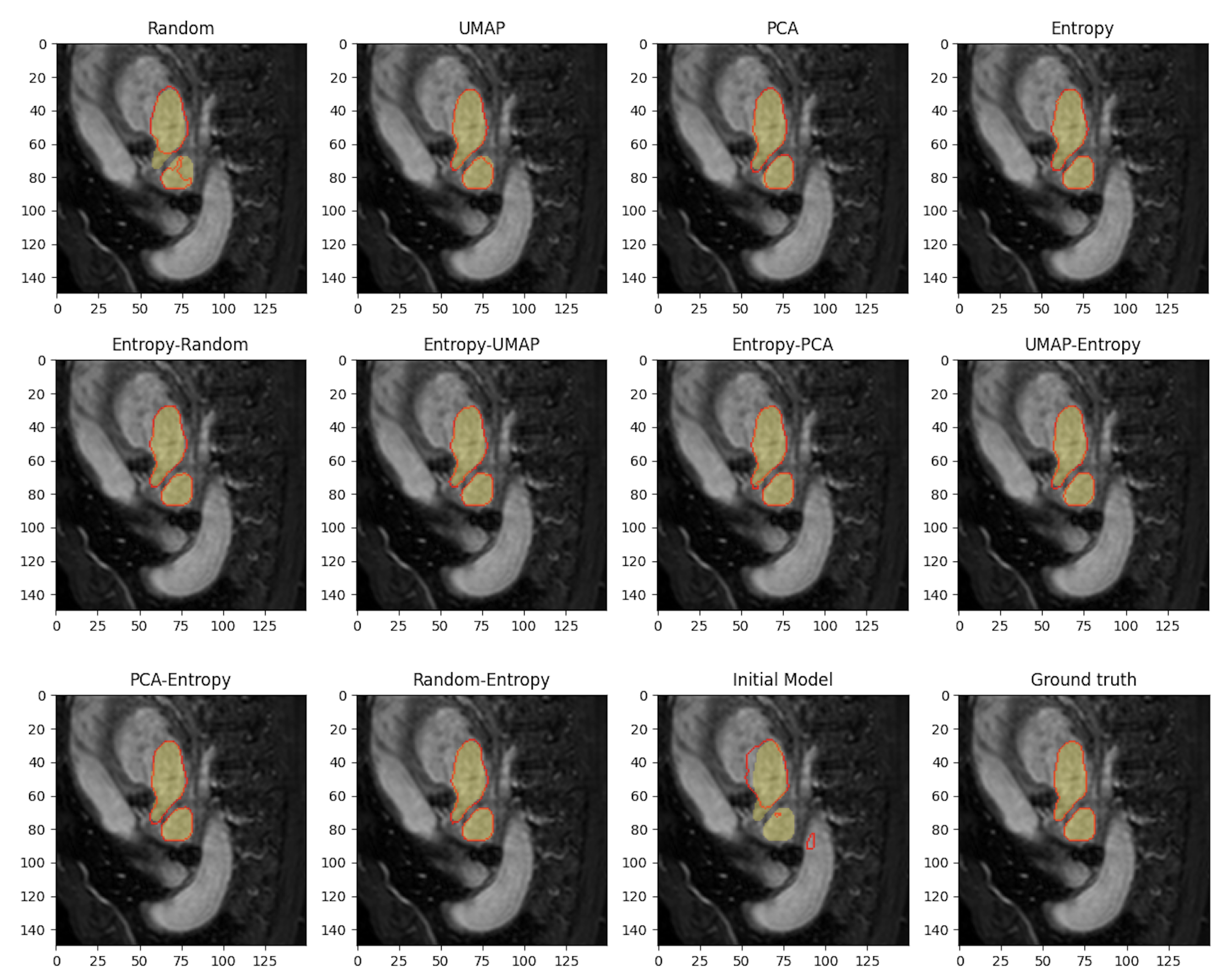}
\caption{An illustrative set of cardiac MRI images (left atrium) with the contours outlined in red for various active learning schemes, overlaid on the ground truth mask in yellow.  }
\label{fig:cardiacR}
\end{figure}

In Figures \ref{fig:prostateR} and \ref{fig:cardiacR}, we furnish some examples of the contoured images across all the active learning methods explored in our work. 

\newpage

\bibliographystyle{agsm}
\bibliography{referenceMore2}

\end{document}